\documentclass[10pt,twocolumn,letterpaper]{article}

\usepackage{iccv}

\usepackage{times}
\usepackage{epsfig}
\usepackage{graphicx}
\usepackage{amsmath}
\usepackage{amssymb}

\usepackage{booktabs}
\usepackage{color}
\usepackage{alltt}
\usepackage{amsfonts}
\usepackage{mathrsfs}
\usepackage{algorithm}
\usepackage{algorithmic}
\usepackage{multicol,multirow}
\usepackage[dvipsnames]{xcolor,colortbl}
\usepackage{makecell}
\usepackage[accsupp]{axessibility}

\usepackage{subcaption}

\usepackage[pagebackref=true,breaklinks=true,colorlinks,bookmarks=false]{hyperref}

\iccvfinalcopy 

\def\iccvPaperID{6875} 

\ificcvfinal\fi

\begin{document}

\title{
NormKD: Normalized Logits for Knowledge Distillation 
}

\author{Zhihao Chi\\
Zhejiang University
\and
Tu Zheng\\
Fabu
\and
Hengjia Li\\
Zhejiang University\\
\and
Zheng Yang\\
Fabu
\and
Boxi Wu\\
Zhejiang University
\and
Binbin Lin\\
Zhejiang University
\and
Deng Cai\\
Zhejiang University
}

\maketitle
\ificcvfinal\thispagestyle{empty}\fi

\begin{abstract}
    Logit based knowledge distillation gets less attention in recent years since feature based methods perform better in most cases. Nevertheless, we find it still has untapped potential when we re-investigate the temperature, which is a crucial hyper-parameter to soften the logit outputs. For most of the previous works, it was set as a fixed value for the entire distillation procedure.
    However, as the logits from different samples are distributed quite variously, it is not feasible to soften all of them to an equal degree by just a single temperature, which may make the previous work transfer the knowledge of each sample inadequately.
    In this paper, we restudy the hyper-parameter temperature and figure out its incapability to distill the knowledge from each sample sufficiently when it is a single value.
    To address this issue, we propose \textbf{Normalized Knowledge Distillation (NormKD)}, with the purpose of customizing the temperature for each sample according to the characteristic of the sample's logit distribution. Compared to the vanilla KD, NormKD barely has extra computation or storage cost but performs significantly better on CIRAR-100 and ImageNet for image classification. Furthermore, NormKD can be easily applied to the other logit based methods and achieve better performance which can be closer to or even better than the feature based method.
\end{abstract}



\begin{figure}
\setlength{\abovecaptionskip}{5pt} 
  \centering
   \includegraphics[width=\linewidth,height=0.95\linewidth]{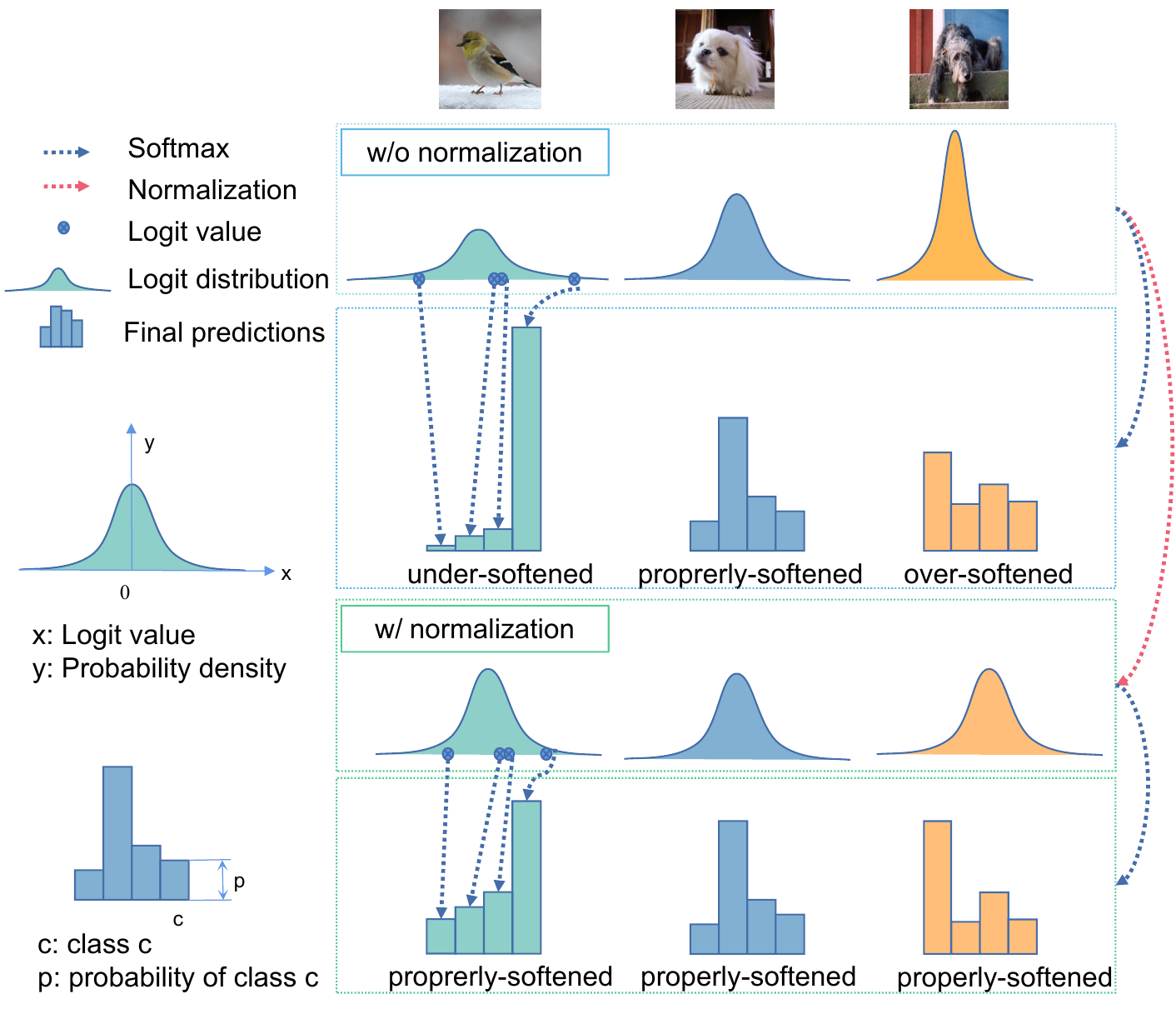}
\setlength{\belowcaptionskip}{-15pt} 
\caption{Illustration of different samples' final predictions with a fixed temperature and normalized temperatures. We call the soft labels which are proper for distillation ``properly softened'' and the ones that are processed by a temperature, which is too large or too small for them, ``over-softened'' and ``under-softened''.
}
\label{fig1}
\end{figure}

\section{Introduction}
Impressive progress has been made in deep neural networks in the last decades, which also promoted the development of computer vision. Tasks such as image classification \cite{vgg,resnet,senet,shufflenetv2}, object detection \cite{faster_rcnn,fpn}, and semantic segmentation \cite{fcn,shelhamer2016fully} have made impressive progress, and can be already deployed into real-life applications in many situations. However, to pursue better performance, the size of neural networks is growing much larger, which prevents these high-performance models to be applied to mobile devices. Knowledge distillation (KD) is an effective technique to address this issue by transferring knowledge from heavy-weighted teacher models to the lighter student.

According to the previous works, we can distill the knowledge from the logit outputs, intermediate feature maps, or relationships among the samples. Logit based KD is the most classic method and the temperature is a crucial hyper-parameter for it. Using a temperature to soften the logit output is an essential step to acquire more ``dark knowledge'' and it is also a part of the reason why this method is called ``distillation''.
However, less attention has been put to the logit based method, since feature based methods dominate the performance in recent years.
A typical recent work is DKD \cite{dkd}, which restudied the potential of logit based KD and achieved profound performance that can be comparable to the state-of-the-art feature based method. 


Nevertheless, most of the previous works set a fixed value as temperature no matter how the various logit outputs distribute, which is actually not suitable. When using a fixed value, there may exist a phenomenon as depicted at the top of Figure \ref{fig1}. We assume the sample in the middle has a suitable soft label, as its predictions are not too close or too far.
For the first sample, which has a confident prediction that is close to a ground-truth label, its soft label would contain inadequate ``dark knowledge'' since the fixed temperature is too small for it. The third sample is less confident for its final predictions are close. The fixed temperature is obviously too large for it, which makes the soft label lose the information of variety and unsuitable for distillation. 
As a consequence, we can find that one temperature has no ability to properly soften the logit output for all samples.
 As the distributions of the different samples differ, one temperature can not soften all the logits to a proper degree equally. And our target is to acquire properly softened labels for all samples as shown at the bottom of Figure \ref{fig1}.

Furthermore, from the experiments we implement in this paper (details in the section \ref{method:mtt}), we can discover that performance is improved apparently when we apply multiple temperatures to the vanilla KD. This phenomenon can also provide us evidence of the incapability of a single temperature, and inspire us that there is something being ignored with the temperature. With the observation of the output distribution and results of the experiment, we can assume that a single temperature is not enough to distill the knowledge adequately from various samples as their outputs are distributed variously.

In this paper, we proposed a novel knowledge distillation approach called Normalized Knowledge Distillation (NormKD), which is simple, with barely extra cost, and can be a complement to the classic logit-based method \cite{kd}. NormKD sets a specific temperature, which is multiple of the standard variance in our method, for each sample according to the distribution of its logit output. We rearrange the logit outputs of various samples into a similar normalized distribution for both teacher and student, which makes different samples get a relatively equal distribution and help the ``dark knowledge'' to be transferred more sufficiently.

Overall, our contributions can be summarized as follows:
\begin{itemize}
\vspace{-0pt}
\setlength{\itemsep}{0pt}
\setlength{\parsep}{0pt}
\setlength{\parskip}{0pt}
\item We restudy the hyper-parameter temperature of vanilla knowledge distillation and figure out that a single temperature is inadequate to properly soften the outputs of different samples.
\item We find that selecting the temperature for each sample by the distribution of sample output can effectively improve the performance of knowledge distillation.
\item We propose a simple and effective logit based knowledge distillation method, which almost introduces no extra cost compared to the classical KD. Moreover, it can be a complement to the other logit based knowledge distillation methods with no conflict and achieve outstanding performance.
\end{itemize}

\section{Related Work}


Knowledge distillation is a technique to train a lighter student model with the supervision of training data and another cumbersome pre-trained teacher model, aiming to minimize the performance gap between the student and teacher. The traditional operation is to align the final predictions by KL-Divergence. For better transferring the ``dark knowledge'', the hyper-parameter temperature is introduced to soften the logits produced by the models.

The main target of knowledge distillation research in recent years is to excavate more information from the teacher model. Since the concept of knowledge distillation was first proposed by Hinton \etal in \cite{kd}, numerous works emerge to raise the performance of students in the last few years. Despite the way of vanilla KD method transferring knowledge by aligning the logit outputs \cite{kd,ba2014deep}, the subsequent works also attempt to distill knowledge by the other part of the neural network. Typically, we can classify KD methods into three categories according to the typical works in recent years, which include logit based KD \cite{kd,ba2014deep,eskd,snapshot,GLD,ipwd,simkd,dkd}, feature based KD \cite{fitnets,ofd,crd,reviewkd,tat,MGD}, and relation based KD \cite{rkd,CIRKD,HKD}. Logit based KD transfers knowledge by minimizing the gap between the student's and teacher's logit outputs using KL-Divergence. Feature based KD concentrates on fitting intermediate layers of student and teacher models. Relation base KD tries to enhance distillation performance by extracting the relationship information among the samples. 

In particular, in recent years, more works turn their interest to the feature based method for its better performance. Fitnet \cite{fitnets} simply aligned one layer of intermediate features. Also, some attempts are made to distill from the feature indirectly, by masking a part of features \cite{MGD}. The subsequent works \cite{at,ofd,crd,tat} introduced other new techniques or design complex new structures, attempting to extract more useful information from the feature layers. Recent work \cite{reviewkd} applied cross-layer distillation and got outstanding performance. Even though the feature based method can achieve state-of-the-art performance in most cases, it would commonly introduce extra computation when calculating the alignment of intermediate layers and extra storage cost when introducing fancy intermediate structures.

As for relation based methods, previous works attempt to build the relationship knowledge among different samples and transfer this kind of knowledge from teacher to student. Typical approach RKD \cite{rkd} constructed distance-wise and angle-wise relationships respectively to transfer the extra relational knowledge.

 Although logit based KD is under-explored over recent years, there are still some impressive works. The recent work \cite{ipwd} indicated the unbalance of knowledge among different samples and devotes to re-weight the importance of samples. SimKD \cite{simkd} reused the teacher model's classifier for student models, achieving good performance. But it is not appropriate to be compared with other logit based methods directly since it changed the structure of the student in the meanwhile. Moreover, DKD \cite{dkd} decoupled the loss function of KL-Divergence into two parts, achieving wonderful performance which can be comparable to the start-of-the-art feature based method.

As we can see, logit based methods still have the potential to achieve an acceptable performance even though they have no ability to transfer the location information as the feature based methods.

In this paper, our key purpose is to advance the logit based KD by customizing the temperature for each sample according to its logit distribution, focusing on excavating the hidden potential of vanilla knowledge distillation.

\section{Method}
\label{method}
In this section, we re-study the effect of temperature on knowledge distillation. We manage to distill the information from teacher models using several different temperatures instead of a fixed one. 
The consequences inspire us that a single temperature is not capable to transfer the knowledge from the logit outputs sufficiently. 
Through the observation of the distribution of final predictions with different temperatures, we propose a simple and efficient logit based distillation approach.

\begin{figure}
\setlength{\abovecaptionskip}{5pt} 
  \centering
	\includegraphics[width=0.99\linewidth,height=0.68\linewidth]{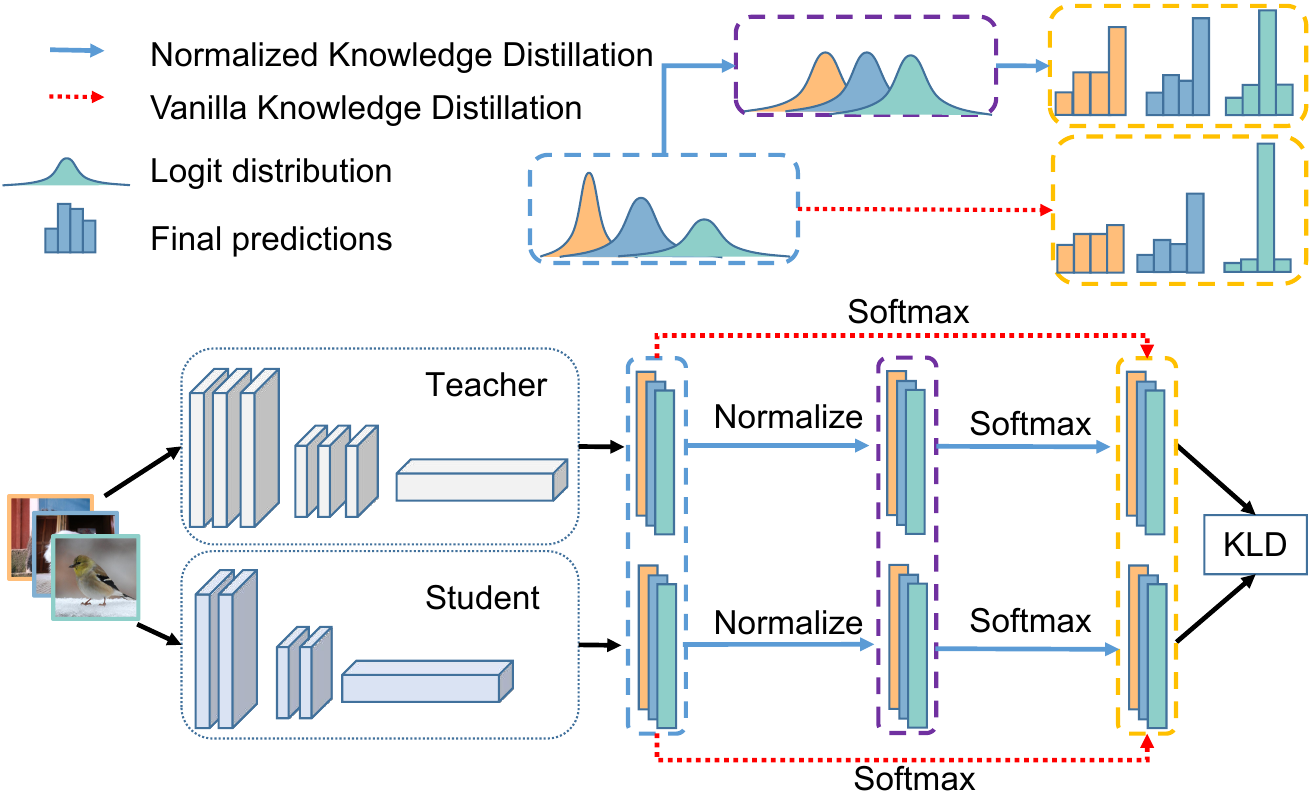}
\setlength{\belowcaptionskip}{-15pt} 
\label{fig2}
\caption{Overview of vanilla KD and NormKD. Compared to vanilla KD, NormKD simply adds a normalize operation before the softmax layers. As shown at the top-right of the Figure, the logit outputs of different samples are distributed differently without normalization and the final predictions are softened unequally. In contrast, the distributions become similar after normalization and the final predictions are softened more equally.}
\end{figure}

\subsection{Vanilla Knowledge Distillation}
The vanilla knowledge distillation method, which is generally applied to classification tasks, only concentrates on aligning the logit outputs between teacher and student. The input images are processed by the non-linear backbone network and turned into a feature map, which is then handled by a linear classifier to acquire the final prediction. Given a training sample $x$ and an one-hot true label $y$, we can get the logit output $\mathbf{z}=[z_1,z_2,...,z_{i},...,z_C] \in \mathbb{R}^{1\times C}$ from the nerual network, where $z_{i}$ is the logit output of the $i$-th class and $C$ is the number of classes. Then we can acquire the final prediction vector $\mathbf{p}=[p_1,p_2,...,p_{i},...,p_C]$ after a softmax layer:
\begin{equation}
    p_{i,T} = \frac{\exp(z_{i}/T)}{\sum_{j=1}^{C} \exp(z_{j}/T)},
\label{pi}
\end{equation}
where T is the hyper-parameter temperature playing the role of label softening, which is set as a fixed value for one student and teacher pair in vanilla KD. And $p_{i,T}$ represents the probability of the class $i$ which is softened by temperature $T$. 

The final loss function consists of cross entropy loss and KL-Divergence loss. The former loss manages the job of classification by the true labels and the latter loss manages the alignment between the student output and the teacher output. The final function can be written as follows:
\begin{equation}
\begin{aligned}
    \mathcal{L}_{KD} &=\alpha \mathcal{L}_{ce}+\beta \mathcal{L}_{kld} \\
                    &=\alpha  CE(\mathbf{p^s_{T=1}},\mathbf{y})+\beta \cdot T^2\cdot KLD(\mathbf{p^s_{T}},\mathbf{p^t_{T}}),
\label{loss_kd}
\end{aligned}
\end{equation}
where $s$ and $t$ represent student and teacher respectively, $\alpha$ and $\beta$ are hyper-parameters to balance the weight of two losses. And the temperature is set to be 1 in the cross entropy loss. $T^2$ is a weight to compensate for the value reduction introduced by temperature in $KLD$.

\subsection{Distill With Multiple Temperature}
\label{method:mtt}
As we know, the previous works generally set the temperature to a fixed value for the entire distillation procedure, which is actually not reasonable. In order to find out whether there is something to be ignored, we introduce multiple temperatures for the vanilla knowledge distillation method. Our method is to soft the logit output $\mathbf{z}$ more than once with different temperature and acquire multiple predictions $[\mathbf{p}_{T=t_1},\mathbf{p}_{T=t_2},...,\mathbf{p}_{T=t_i},...,\mathbf{p}_{T=t_k}]$. Then, we simply calculate the average value of these predictions by 
\begin{equation}
\begin{aligned}
    \bar{\mathbf{p}} = \frac{1}{k} \cdot \sum_{i=1}^k \mathbf{p}_{T=t_i} ,
\label{t_mean}
\end{aligned}
\end{equation}
where $k$ is the number of temperatures we use, and $t_i$ is the value of temperature we select. Then we can rewrite the $\mathcal{L}_{kld}$ as:
\begin{equation}
    \mathcal{L}_{kld} = T_{mul}^2\cdot KLD(\mathbf{\bar p^s},\mathbf{\bar p^t}),
\label{defpi}
\end{equation}
where $T_{mul}^2$ is the compensation weight for reduction introduced by multiple temperatures. We choose the $T_{mul}$ as max value among $[t_1,t_2,...,t_i,...,t_k]$ empirically in the experiment.

In this way, the student is supposed to distill knowledge simultaneously from the soft labels which are softened to different degrees. We implement this method using ResNet \cite{resnet} as training model. From the results in Table~\ref{tab:multi_t}, we can find that the performance is improved quite obviously when we choose an appropriate temperature set. When we choose 14 temperatures, the best result surpasses some effective KD methods. In contrast, when we use a single temperature, as we select 1,2,4, and 8 respectively in Table \ref{tab:multi_t}, can not make obvious progress. 
In addition, the performance of distillation is positively correlated with the number of temperatures in a suitable range. When we use too many temperatures, the accuracy starts to decay. 

Even though this method performs efficiently in our experiment, it is tedious to figure out which temperature may have a positive influence on a specific model, how many temperatures should we select to achieve the best performance, and how to set an appropriate $T_{mul}$ for various temperature after averaging. As a consequence, this method is difficult to be applied generally. But the ascending performance still prompts us that using a single temperature for knowledge distillation is inadequate, pushing us to explore the reason further.

\begin{table}[h]
\setlength{\belowcaptionskip}{-10pt} 
\center
\begin{small}
\begin{tabular}{ccccc}
\multicolumn{1}{c|}{temperature set}          & top-1 & $\Delta$ \\ \hline 
\multicolumn{1}{c|}{$[1]$}                   & 73.33     & -    \\
\multicolumn{1}{c|}{$[2]$}                   & 73.17     & -0.16    \\
\multicolumn{1}{c|}{$[4]$}                   & 73.25     & -0.08    \\
\multicolumn{1}{c|}{$[8]$}                   & 73.55     & +0.22   \\
\multicolumn{1}{c|}{$[1,2]$}                 & 73.37     & +0.04   \\ 
\multicolumn{1}{c|}{$[1,2,4]$}               & 74.78     & +1.45    \\ 
\multicolumn{1}{c|}{$[1,2,4,6]$}             & 75.47     & +2.14     \\ 
\multicolumn{1}{c|}{$[1,2,4,6,8]$}           & 75.70     & +2.37    \\ 
\multicolumn{1}{c|}{$[1,2,3,...,12,13,14]$}  & 76.33     & +3.00     \\ 
\multicolumn{1}{c|}{$[1,2,3,...,12,13,14,15,16]$} & 75.58     & +2.25 \\ \hline
\multicolumn{5}{c}{\textit{ResNet32$\times$4 as the teacher,ResNet8$\times$4 as the student}} \\ \hline
\\
\end{tabular}
\end{small}
\vspace{-8pt} 
\caption{
	Top-1 accuracy(\%) on the CIFAR-100 validation set. The temperature set contains the temperatures we used and $\Delta$ represents the performance improvement over the baseline set [1].
	}
\label{tab:multi_t}
\end{table}

\subsection{Normalized Knowledge Distillation}
\begin{table*}[h]
\setlength{\belowcaptionskip}{-10pt} 
\center
\begin{small}
\begin{tabular}{cc|cccccc}
\multirow{4}{*}{\begin{tabular}[c]{@{}c@{}}distillation \\ manner\end{tabular}} & \multirow{2}{*}{teacher}  & ResNet56 & ResNet110 & ResNet32$\times$4 & WRN-40-2 & WRN-40-2 & VGG13 \\
&      & 72.34      & 74.31       & 79.42        & 75.61      & 75.61      & 74.64   \\
& \multirow{2}{*}{student}  & ResNet20 & ResNet32 & ResNet8$\times$4  & WRN-16-2 & WRN-40-1 & VGG8  \\
& \space     & 69.06      & 71.14       & 72.50        & 73.26      & 71.98      & 70.36   \\ \Xhline{3\arrayrulewidth} 
\multirow{5}{*}{features}
& FitNet \cite{fitnets}   & 69.21      & 71.06       & 73.50        & 73.58      & 72.24      & 71.02   \\
& RKD \cite{rkd}      & 69.61      & 71.82       & 71.90        & 73.35      & 72.22      & 71.48   \\
& CRD \cite{crd}      & 71.16      & 73.48       & 75.51        & 75.48      & 74.14      & 73.94   \\
& OFD \cite{ofd}      & 70.98      & 73.23       & 74.95        & 75.24      & 74.33      & 73.95   \\
& ReviewKD \cite{reviewkd} & 71.89      & 73.89       & 75.63        & 76.12      & \textbf{75.09}      & \textbf{74.84}   \\ \hline 
\multirow{5}{*}{logits}                                                         
& KD \cite{kd}       & 70.66      & 73.08       & 73.33        & 74.92      & 73.54      & 72.98   \\
& DKD \cite{kd}       & \textbf{71.97}      & \textbf{74.11}       & 76.32        & 76.24      & 74.81      & 74.68 \\
& \textbf{NormKD}   &71.40      &73.91         &\textbf{76.57}         & \textbf{76.40}       &74.84      &74.45 \\

& $\Delta_{1}$      & \textcolor{ForestGreen}{+0.74}      & \textcolor{ForestGreen}{+0.83}       & \textcolor{ForestGreen}{+3.24}        & \textcolor{ForestGreen}{+1.48}      & \textcolor{ForestGreen}{+1.30}      & \textcolor{ForestGreen}{+1.47} \\

& \textbf{DKD+NormKD}   &\textit{\textbf{71.99}}  &\textit{\textbf{74.28}}  &\textit{\textbf{77.33}} &\textit{\textbf{76.59}} &\textit{\textbf{75.36}} &\textit{\textbf{75.19}}\\
& $\Delta_{2}$  &\textcolor{ForestGreen}{+0.02}  &\textcolor{ForestGreen}{+0.17}  &\textcolor{ForestGreen}{+1.01}  &\textcolor{ForestGreen}{+0.35}  &\textcolor{ForestGreen}{+0.55}  &\textcolor{ForestGreen}{+0.51}  \\

\end{tabular}
\end{small}
\vspace{-8pt} 
\caption{\textbf{Results on the CIFAR-100 validation.} Teachers and students are in the \textbf{same} architectures. $\Delta_{1}$ represents the performance improvement of NormKD over the classical KD, and $\Delta_{2}$ represents the performance improvement of DKD+NormKD over the DKD. All results are the average over 3 trials.}
\label{tab:cifar}
\end{table*}

As mentioned in the above section, only one temperature has no capability to extract all the knowledge that students can learn and multiple temperatures lead to a better result. As we know, a higher value temperature makes the logit output softer while a lower value affects the output oppositely. The only difference between the multiple temperatures method and the vanilla KD is that it distills the knowledge from the variously softened labels simultaneously. With the observation of teacher models' final predictions of different samples with different temperatures, we also discover a single temperature can not soften the logit output equally for all samples. A part of them would be over-softened or under-softened, inducing themselves unsuitable for distillation. And this could be the reason for the results we get from section \ref{method:mtt}.

\begin{figure}
\setlength{\abovecaptionskip}{5pt} 
  \centering
   \includegraphics[width=\linewidth,height=0.4\linewidth]{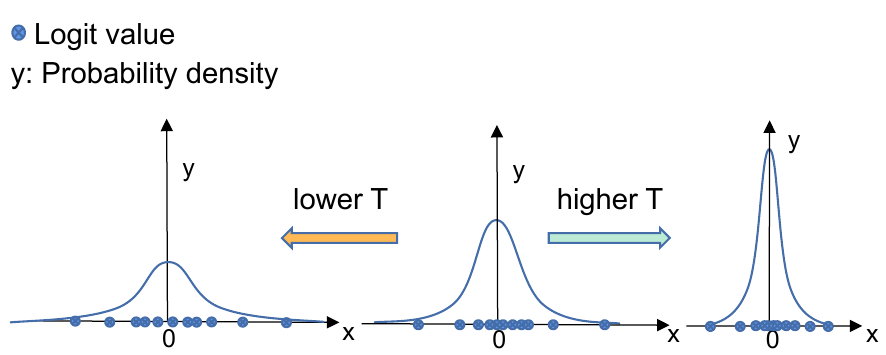}
\setlength{\belowcaptionskip}{-15pt} 
\caption{Illustration of logit distributions when the temperature changes for one sample. T is the temperature. The logits are distributed closer when the temperature grows and more separately when the temperature declines. The distribution is similar to normal distribution. }
\label{fig3}
\end{figure}

Therefore, selecting a specific value as the temperature for each sample is a feasible idea. 
With the aim to soften the output of all the samples equally, we attempt to find a value that is related to the characteristic of the sample itself. 
By observing the distribution of a sample's logit output when the temperature is a fixed value in the middle of Figure \ref{fig3}, we can easily discover that most values are located around zero and the values far from the middle are much less, which is similar to normal distribution. The values are distributed more separately when the temperature declines and become closer when the temperature grows. Therefore, it is appropriate to approximately regard the logits distribution as a normal distribution with location parameter $\mu$ and scale parameter $\sigma$. And temperature plays a role to modify the scale of distribution, which is similar to the $\sigma$ of normal distribution.

As mentioned above, when we approximately regard the distribution as a normal distribution, it would be easy to rescale different distributions into a standard one. Then we can find our aim to select a specific temperature for a sample is achieved. Our normalized knowledge distillation is simple, only adding a normalization operation before the softmax layers, which can be described as follow:
\begin{equation}
\begin{aligned}
    \widetilde p_{i} &= \frac{\exp((z_{i}-\mu)/\sigma)}{\sum_{j=1}^{C} \exp((z_{j}-\mu)/\sigma)}, \\
               &= \frac{\exp(z_{i}/\sigma)/\exp(\mu/\sigma)}{\sum_{j=1}^{C}(\exp(z_{j}/\sigma)/\exp(\mu/\sigma))}, \\
               &= \frac{\exp(z_{i}/\sigma)/\exp(\mu/\sigma)}{(\sum_{j=1}^{C} \exp(z_{j}/\sigma))/\exp(\mu/\sigma)}, \\
               &= \frac{\exp(z_{i}/\sigma)}{\sum_{j=1}^{C} \exp(z_{j}/\sigma)},
\end{aligned}
\label{norm_p}
\end{equation}
where $C$ is the number of classes, $\mu$ and $\sigma$ are the mean and standard variance of a sample's logit output $[z_1,z_2,...,z_{i},...,z_C]$ respectively, $\widetilde p_{i}$ represents the final prediction after normalization for class $i$. And we can find the mean value $\mu$ takes no effect on the final $\widetilde p_{i}$. Thus, the only thing we need to do is to calculate the standard variance of the logit and divide it.

Compared to Eq \eqref{pi}, the only difference is the new $\widetilde p_{i}$ uses $\sigma$ to replace the original temperature. However, the standard distribution may not be the most suitable for the models, we still need a hyper-parameter to scale the distribution again to search for a satisfactory result. Thus, we introduce another scaling parameter $T_{norm}$ for Eq \eqref{norm_p}, and reformulate $\widetilde p_{i}$ as follow:
\begin{equation}
\begin{aligned}
   \widetilde p_{i} &= \frac{\exp(z_{i}/(\sigma \cdot T_{norm}))}{\sum_{j=1}^{C} \exp(z_{j}/(\sigma \cdot T_{norm}))}.
\end{aligned}
\label{norm_p_final}
\end{equation}

Finally, we can rewrite the $\mathcal{L}_{kld}$ as:
\begin{equation}
    \mathcal{L}_{kld} =\frac{1}{N} \cdot \sum_{i=1}^{N} ( (T_{norm}\cdot \sigma_{i}^t)^2\cdot KLD(\mathbf{\widetilde p^s_i},\mathbf{\widetilde p^t_i})),
\label{kld_final}
\end{equation}
where $N$ is the number of samples, $\sigma_{i}^t$ is the standard variance of the teacher model's logit output for sample $i$ and $\mathbf{\widetilde p_i}$ is the normalized final prediction of sample $i$. We apply the normalization for both teacher and student logit. Since the pre-trained teacher has a constant output for each sample and the student's output varies during the training process, we set the square of the teacher's final temperature $(T_{norm}\cdot \sigma_{i}^t)^2$ as the compensation weight for each sample. 

\begin{algorithm}[t]
    \caption{Pseudo code of NormKD in a PyTorch-like style.}
    \label{algo:norm}
    \footnotesize
    \begin{alltt}
    \color{ForestGreen}
# z_stu: student output logits
# z_tea: teacher output logits
# sigma_stu: standard variance of student logit 
# sigma_tea: standard variance of teacher logit 
# T_s: final temperature for student 
# T_t: final temperature for teacher 
# p_stu: final prediction for student 
# p_tea: final prediction for teacher 
# T_norm: hyper-parameter of NormKD
\end{alltt}
\vspace{-15pt}
\begin{alltt}
\color{ForestGreen}# calculate the std of each sample's logit \color{Black}
sigma_stu = z_stu.std(1,keepdim=True)
sigma_tea = z_tea.std(1,keepdim=True)

\color{ForestGreen}# use T_norm to re-scale the final temperatures \color{Black}
T_s = sigma_stu * T_norm
T_t = sigma_tea * T_norm

p_stu = F.softmax(z_stu / T_s)
p_tea = F.softmax(z_tea / T_t)

kld = F.kl_div(log(p_stu), p_tea)
loss_normkd
    =(kld.sum(1,keepdim=True))*(T_t**2)).mean()
\end{alltt}
\end{algorithm}

To sum up, our NormKD replaces the temperature with the scaled standard variance of each sample's logit output. The process is similar to normalization without subtracting the average value, which makes each sample's output to be closer in distribution. The only computation introduced by NormKD is to calculate the standard variance for every logit output. To some extent, it can solve the inequality problem introduced by a single temperature and improve the performance of distillation profoundly. Compared to the vanilla KD method, NormKD has nearly the same computation cost, the same number of hyper-parameter, and much better performance. It can be a perfect complement to the vanilla KD method. Furthermore, it is also easy to be combined with other distillation methods for its simplicity.

Algorithm \ref{algo:norm} provides the pseudo-code of NormKD in a PyTorch-like  \cite{torch} style which is simple and similar to the vanilla KD.

\begin{table*}[th]
\setlength{\belowcaptionskip}{-10pt} 
\center
\begin{small}
\begin{tabular}{cc|ccccc}
\multirow{4}{*}{\begin{tabular}[c]{@{}c@{}}distillation \\ manner\end{tabular}} & \multirow{2}{*}{teacher}  & ResNet32$\times$4 & WRN-40-2 & VGG13 & ResNet50 & ResNet32$\times$4 \\
& \space     & 79.42      & 75.61       & 74.64        & 79.34      & 79.42         \\
& \multirow{2}{*}{student}  & ShuffleNet-V1 & ShuffleNet-V1 & MobileNet-V2  & MobileNet-V2 & ShuffleNet-V2  \\
& \space     & 70.50      & 70.50       & 64.60        & 64.60      & 71.82         \\ \Xhline{3\arrayrulewidth} 
\multirow{5}{*}{features}
& FitNet \cite{fitnets}   & 73.59      & 73.73       & 64.14        & 63.16      & 73.54         \\
& RKD \cite{rkd}      & 72.28      & 72.21       & 64.52        & 64.43      & 73.21         \\
& CRD \cite{crd}      & 75.11      & 76.05       & 69.73        & 69.11      & 75.65         \\
& OFD \cite{ofd}      & 75.98      & 75.85       & 69.48        & 69.04      & 76.82         \\
& ReviewKD \cite{reviewkd} & \textbf{77.45}      & \textbf{77.14}       & \textbf{70.37}        & 69.89      & \textbf{77.78}         \\ \hline 
\multirow{5}{*}{logits}                                                         
& KD \cite{kd}       & 74.07      & 74.83       & 67.37        & 67.35               & 74.45         \\
& DKD \cite{dkd}      & 76.45      & 76.70       & 69.71       & \textbf{70.35}      & 77.07       \\
& \textbf{NormKD}    &75.62       &76.62       &69.53          &69.57                  &76.01  \\

& $\Delta_{1}$      & \textcolor{ForestGreen}{+1.55}  & \textcolor{ForestGreen}{+1.79}  & \textcolor{ForestGreen}{+2.16}        & \textcolor{ForestGreen}{+2.22}      & \textcolor{ForestGreen}{+1.56} \\

& \textbf{DKD+NormKD}  &76.81  &76.99 &70.22 &\textit{\textbf{70.91}}  &77.13  \\
& $\Delta_{2}$  &\textcolor{ForestGreen}{+0.36} &\textcolor{ForestGreen}{+0.29}  &\textcolor{ForestGreen}{+0.51} &\textcolor{ForestGreen}{+0.56}  &\textcolor{ForestGreen}{+0.05} \\

\end{tabular}
\end{small}
\\
\vspace{-8pt} 
\caption{\textbf{Results on the CIFAR-100 validation.} Teachers and students are in \textbf{different} architectures. $\Delta_{1}$ represents the performance improvement of NormKD over the classical KD, and $\Delta_{2}$ represents the performance improvement of DKD+NormKD over the DKD. All results are the average over 3 trials.}
\label{tab:cifar2}
\end{table*}
\section{Experiments}

In this section, we apply our NormKD to various teacher and student distillation pairs and conduct the training process on several standard benchmark datasets. Firstly, we compare its performance with the previous approaches. Then, we try to combine our method with the previous logit based method, since NormKD is simple and easy to be applied.

\subsection{Datasets and baselines.}

We mainly apply our method on classification tasks with two benchmark datasets as follows:

{\textbf{CIFAR-100} \cite{cifar}} is a tiny labeled dataset with 60k images, consisting of 50k images for training and 10k images for validation. The dataset is popular for image classification tasks, as it contains 100 categories.

{\textbf{ImageNet} \cite{imagenet} is a well-known large dataset for image classification. It contains 1000 classes and more than 1.2 million images in total. There are 1.28 million images for training and 50k images for validation.

Despite comparison with the vanilla method, we also compare our method with the other previous works including FitNet \cite{fitnets}, RKD \cite{rkd}, CRD \cite{crd}, AT \cite{at}, OFD \cite{ofd}, ReviewKD \cite{reviewkd} and DKD \cite{dkd}. VGGNet \cite{vgg}, ResNet \cite{resnet}, WideResNet \cite{wrn}, ShuffleNet  \cite{shufflenetv1,shufflenetv2} and MobileNet-V2 \cite{mobilenetv2} are selected as teacher or student models.

\subsection{Main Results}

\begin{table*}[h]
\center
\begin{small}
\begin{tabular}{ccc|cccc|cccc}
\multicolumn{3}{c|}{distillation manner} & \multicolumn{4}{c|}{features}    & \multicolumn{4}{c}{logits} \\ \Xhline{\arrayrulewidth}
           & teacher   & student     & AT \cite{at}  & OFD \cite{ofd}  & CRD \cite{crd}   & ReviewKD \cite{reviewkd}  & KD \cite{kd}            & DKD \cite{dkd}        &\textbf{NormKD}   &\textbf{NormKD+DKD}  \\ \hline
top-1      & 73.31     & 69.75       & 70.69        & 70.81          & 71.17           & 71.61                 & 70.66                & 71.70      &71.56   &\textbf{72.03}  \\
top-5      & 91.42     & 89.07       & 90.01        & 89.98          & 90.13           & 90.51       & 89.88                 & 90.41       &90.47    &\textbf{90.64} \\ 
\end{tabular}
\vspace{-8pt} 
\caption{\textbf{Top-1 and top-5 accuracy~(\%) on the ImageNet validation.} We set \textbf{ResNet-34} as the teacher and \textbf{ResNet-18} as the student. All results are the average over 3 trials.}
\label{tab:imgaenet1}

\setlength{\belowcaptionskip}{-10pt} 
\begin{tabular}{ccc|cccc|cccc}
\multicolumn{3}{c|}{distillation manner} & \multicolumn{4}{c|}{features}    & \multicolumn{4}{c}{logits} \\ \Xhline{3\arrayrulewidth}
           & teacher      & student     & AT \cite{at}    & OFD \cite{ofd}    & CRD \cite{crd}   & ReviewKD \cite{reviewkd} & KD \cite{kd}              & DKD \cite{dkd}       &\textbf{NormKD}   &\textbf{NormKD+DKD}\\ \hline
top-1      & 76.16        & 68.87       & 69.56         & 71.25             & 71.37           & 72.56       & 68.58                   & 72.05            &72.12    &\textbf{72.79}      \\
top-5      & 92.86        & 88.76       & 89.33         & 90.34             & 90.41           & 91.00                & 88.98                 &  91.05   &90.86          &\textbf{91.08}\\
\end{tabular}
\end{small}
\vspace{-8pt} 
\caption{\textbf{Top-1 and top-5 accuracy~(\%) on the ImageNet validation.} We set \textbf{ResNet-50} as the teacher and \textbf{MobileNet-V1} as the student. All results are the average over 3 trials.}
\label{tab:imgaenet2}
\end{table*}

{\textbf{Image Classification on CIFAR-100.}} We discuss experimental results on CIFAR-100 to evaluate the effect our NormKD. The validation accuracy is reported in Table \ref{tab:cifar} and Table \ref{tab:cifar2} respectively. Table \ref{tab:cifar} shows the results of teachers and students with the same network architectures and Table \ref{tab:cifar2} shows the results of teachers and students with different architectures.

From Table \ref{tab:cifar} and Table \ref{tab:cifar2}, we can see that NormKD can perform better than DKD and ReviewKD on the models ResNet32×4/ResNet8×4 and WRN-40-2/WRN-16-2. 
Compared to classical KD, it achieves obviously improvement in all the models no matter what the architectures of the models. Since NormKD is so simple and introduces nearly no extra cost, the improvement is quite profound. Moreover, NormKD can be combined with other logit based methods smoothly for its simplicity. And the results on the bottom of Table \ref{tab:cifar} and Table \ref{tab:cifar2} show that DKD apparently performs better when we apply NormKD on it. Especially, the accuracy is improved by more than 1.0\% on the model ResNet32×4/ResNet8×4. And the final results are further close to the feature based ReviewKD in the models with different structures and better in all models with the same structures, which reveals the hidden potential of logit based method.

{\textbf{Image Classification on ImageNet.}} Table \ref{tab:imgaenet1} and Table \ref{tab:imgaenet2} contain the results on ImageNet, which can prove the effectiveness of NormKD further. Especially, on the model set of ResNet-50/MobileNet-V1, NormKD performs better than DKD in the top-1 accuracy, but worse in the top-5 accuracy. Overall, similar to the performance on CIFAR-100, NormKD can improve the performance of vanilla KD quite obviously. When NormKD is combined with the effective logit based method DKD, it can not only improve the performance of DKD significantly, which achieves around 0.3\% and 0.7\% improvement respectively but also exceeds the performance of ReviewKD on both models.

\subsection{Training details}
We adopt the training procedure of DKD \cite{dkd}, and apply NormKD to the models with the same structures and different structures for comparison. For CIFAR-100, we set the mini-batch size 64 and the weight decay $5 \times 10^{-4}$. And the initial learning rate is set to 0.05 for all models except for the  MobileNet \cite{mobilenetv2}/ShuffleNet \cite{shufflenetv1,shufflenetv2} series architectures, for which the initial learning rate is set to 0.01. We set the training epoch as 240 in total while the learning rate decays with the decay rate 0.1 at the 150th, 180th, and 210th epochs. For ImageNet, the minibatch is set to 512 and the weight decay is set to $1 \times 10^{-4}$. The total epoch is set as 100, while the initial learning rate is set to 0.2 and then divided by 10 at the 30th, 60th, and 90th training epochs. For all the datasets, we adopt SGD optimizer with 0.9 Nesterov momentum. 

As for the hyper-parameter, $T_{norm}$ is set to 2.0 for CIFAR-100. For ImageNet, we set it to 1.0. $\alpha$ and $\beta$, which is to balance the cross entropy loss and KL-Divergence loss, are set to 0.1 and 0.9 respectively for CIFAR-100. For ImageNet, $\alpha$ and $\beta$ are set to 0.5 and 0.5.
And all the cases are trained on GPU NVIDIA 3080TI. We use 1 GPU for CIFAR-100 and 4 for ImageNet. All the results are average of over 3 trials. 

\subsection{Ablation}

\textbf{Hyper-parameter $T_{norm}$}. When using the standard variance to replace the hyper-parameter temperature of the classical KD, NormKD introduces another hyper-parameter $T_{norm}$. Table \ref{tab:Tnorm} demonstrates the results with different values of $T_{norm}$. We can get the best performance when setting $T_{norm}$ to 1.5 on the ResNet32$\times$4/ResNet8$\times$4 for CIFAR-100, and when it is set to 2.0, it also has a close performance. With the experimental experience on various models, it would be feasible to set $T_{norm}$ to 2.0 in most cases. For ImageNet, as shown in Table \ref{tab:Tnorm_imagenet}, we get the best result on the ResNet-50/MobileNet-V1 when $T_{norm}$ is set to 1.0. 

\begin{table}[h]
\setlength{\belowcaptionskip}{-10pt} 
\center


\begin{small}
\begin{tabular}{ccccccc}
\multicolumn{1}{c|}{$T_{norm}$}     &0.75 &1.0 &1.5 &2.0 &3.0 &4.0   \\ \hline 
\multicolumn{1}{c|}{top-1}        &75.87  & 76.26 & \textbf{76.67} & 76.57 & 76.17 & 75.02  \\
\end{tabular}
\end{small}
\vspace{-8pt} 
\caption{
	Ablation of $T_{norm}$. Results with different $T_{norm}$ on the CIFAR-100 validation set. ResNet32$\times$4 as the teacher, ResNet8$\times$4 as the student.
 }
\label{tab:Tnorm}
\end{table}

\begin{table}[h]
\setlength{\belowcaptionskip}{-10pt} 
\center

\begin{small}
\begin{tabular}{ccccccc}
\multicolumn{1}{c|}{$T_{norm}$}     &0.5 &1.0 &1.5 &2.0  \\ \hline 
\multicolumn{1}{c|}{top-1}        &71.44  & \textbf{72.22} &72.14  & 71.91   \\
\multicolumn{1}{c|}{top-5}        &90.39   &\textbf{90.93} &90.82  &90.85
\end{tabular}
\end{small}

\vspace{-8pt} 
\caption{
	Ablation of $T_{norm}$. Results with different $T_{norm}$ on the ImageNet validation set. ResNet-50 as the teacher, MobileNet-V1 as the student.
 }
\label{tab:Tnorm_imagenet}
\end{table}

\begin{table}[h]
\setlength{\belowcaptionskip}{-10pt} 
\center
\begin{small}
\begin{tabular}{ccccc}
\multicolumn{1}{c|}{$T$}          & top-1   \\ \hline 
\multicolumn{1}{c|}{$V_{max}$}                  &76.23    \\
\multicolumn{1}{c|}{$V_{max}-V_{min}$}          &76.16        \\
\\
\end{tabular}
\end{small}
\vspace{-8pt} 
\caption{
	Results with different temperatures on the CIFAR-100 validation set. ResNet32$\times$4 as the teacher, ResNet8$\times$4 as the student.
 }
 \label{tab:max_min}
\end{table}

\textbf{Other temperatures.} Except for the standard variance used in NormKD, we also attempt to find other characteristics of logit distribution to replace the original temperature. We denote the maximum value of a logit output as $V_{max}$ and the minimum value as $V_{min}$. In our experiment, we simply use $V_{max}$ and $V_{max}-V_{min}$ to replace the temperature respectively. And we also use a hyper-parameter $T_{v}$, which is similar to $T_{norm}$ in NormKD, to rescale the $V_{max}$ and $V_{max}-V_{min}$. The results in the table \ref{tab:max_min} indicate the effectiveness of these two methods. Although they can't perform as well as NormKD, both of them can improve the performance of vanilla KD profoundly. These results can also prove the effectiveness of our method to find temperature according to the distribution of logit. 

\textbf{Training efficiency.} 
As NormKD only adds a normalize operation before the softmax layer compared to vanilla KD, it has a similar training efficiency. We test the training time of several KD methods on Table \ref{tab:efficiency} with the setting of ResNet32$\times$4/ResNet8$\times$4. NormKD costs the training time which is close to the vanilla KD. Meanwhile, it introduces no extra parameters while achieving impressive performance.
\begin{table}[h]
\setlength{\belowcaptionskip}{-10pt} 
\center
\begin{small}
\begin{tabular}{ccccccc}
\multicolumn{1}{c|}{ }                      &top-1    &time(ms)  &params   \\ \hline 
\multicolumn{1}{c|}{KD \cite{kd}}            &73.33    &7.1      &0 \\
\multicolumn{1}{c|}{RKD \cite{rkd}}          &71.90    &21      &0  \\
\multicolumn{1}{c|}{FitNet \cite{fitnets}}   &73.50    &8.0      &16.8K  \\
\multicolumn{1}{c|}{OFD\cite{ofd}}           &74.95    &25      &86.9K   \\
\multicolumn{1}{c|}{CRD \cite{crd}}          &75.51    &24      &12.3M   \\
\multicolumn{1}{c|}{ReviewKD \cite{reviewkd}}&75.63    &11      &1.8M   \\
\multicolumn{1}{c|}{DKD \cite{dkd}}          &76.32    &7.5      &0  \\ \hline
\multicolumn{1}{c|}{NormKD}                  &76.57    &7.2      &0   \\ 
\end{tabular}
\end{small}
\vspace{-8pt} 
\caption{
	Training time (per batch) and parameter numbers on CIRAR-100.
 }
\label{tab:efficiency}
\end{table}

\subsection{Visualization}
We present the visualization by t-SNE with the setting of ResNet32×4 as teacher and ResNet8×4 as student on CIFAR-100. We can observe that the effect is improved obviously when our NormKD is applied to vanilla KD and DKD, as the representations on Figure \ref{tsne:normkd} and Figure \ref{tsne:dkd_normkd} are separated more apparently. And the performance of NormKD is also better than DKD on this model pair. 

In addition, we visualize the correlation matrices of student and teacher's logits in Figure \ref{co_re} with the setting of ResNet32×4 as teacher and ResNet8×4 as student on CIFAR-100. The deeper color means the logits of student and teacher are more different. We can find the difference between student and teacher's logits becomes larger when we apply NormKD to vanilla KD. It is because NormKD doesn't align the logits directly but after normalization. When we compare the correlation matrices of the normalized logits between vanilla KD and NormKD, we can discover that NormKD helps the student to acquire more similar normalized logits, which makes it achieve better performance.
\begin{figure}
\setlength{\abovecaptionskip}{5pt} 
  \centering
  \begin{subfigure}{0.49\linewidth}
	\includegraphics[width=0.9\linewidth,height=0.9\linewidth]{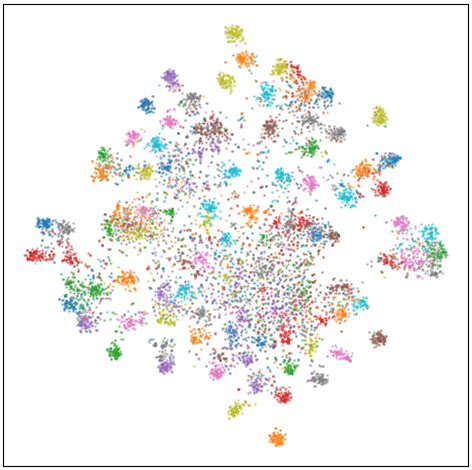}
    \caption{KD}
    \label{tsne:kd}
  \end{subfigure}
  \begin{subfigure}{0.49\linewidth}
	\includegraphics[width=0.9\linewidth,height=0.9\linewidth]{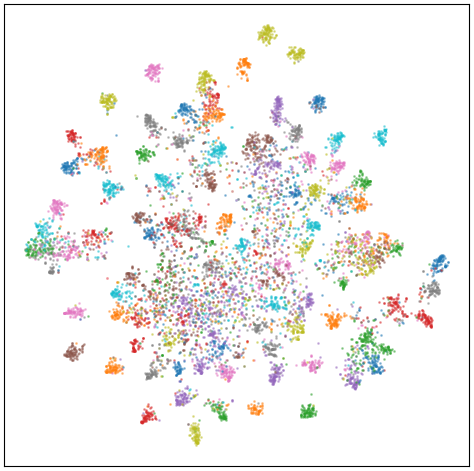}
    \caption{DKD}
    \label{tsne:dkd}
  \end{subfigure}
    \begin{subfigure}{0.49\linewidth}
	\includegraphics[width=0.9\linewidth,height=0.9\linewidth]{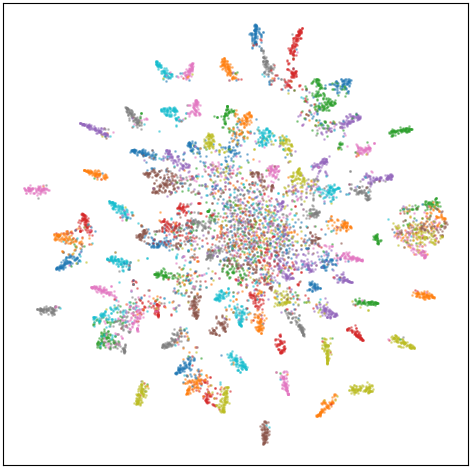}
    \caption{NormKD}
    \label{tsne:normkd}
  \end{subfigure}
    \begin{subfigure}{0.49\linewidth}
	\includegraphics[width=0.9\linewidth,height=0.9\linewidth]{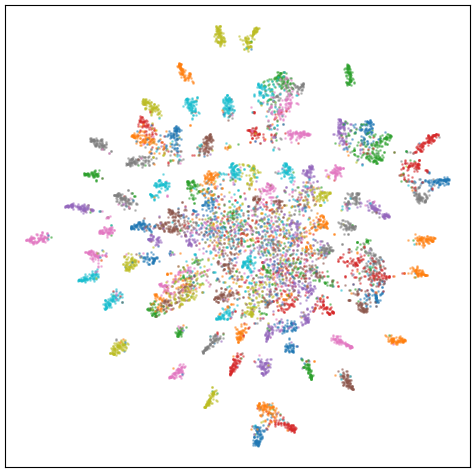}
    \caption{DKD+NormKD}
    \label{tsne:dkd_normkd}
  \end{subfigure}
\setlength{\belowcaptionskip}{-15pt} 
\caption{t-SNE visualization for different methods.}
\label{tsne}
\end{figure}


\begin{figure}
\setlength{\abovecaptionskip}{5pt} 
  \centering
  \begin{subfigure}{0.49\linewidth}
	\includegraphics[width=0.99\linewidth]{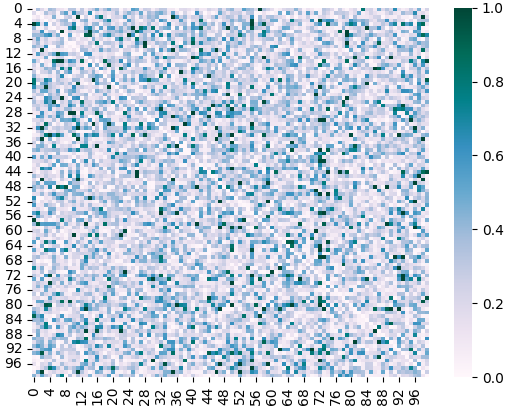}
    \caption{KD}
    \label{f_kd}
  \end{subfigure}
  \begin{subfigure}{0.49\linewidth}
	\includegraphics[width=0.99\linewidth]{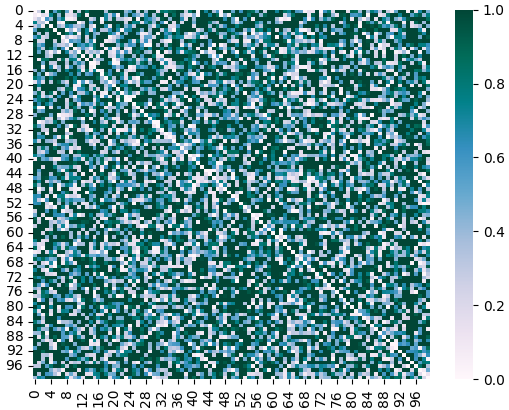}
    \caption{NormKD}
    \label{dkd}
  \end{subfigure}
    \begin{subfigure}{0.49\linewidth}
	\includegraphics[width=0.99\linewidth]{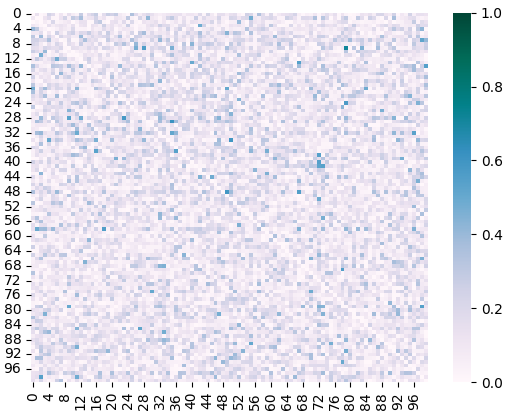}
    \caption{KD*}
    \label{normkd}
  \end{subfigure}
    \begin{subfigure}{0.49\linewidth}
	\includegraphics[width=0.99\linewidth]{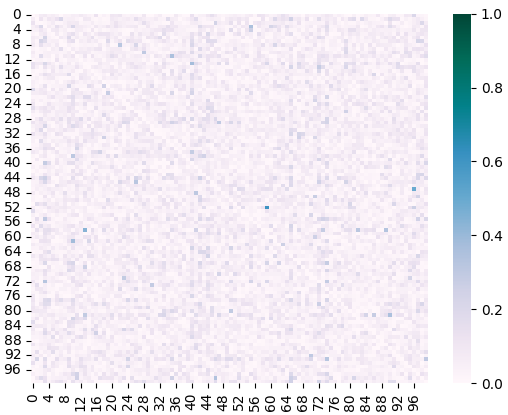}
    \caption{NormKD*}
    \label{dkd_normkd}
  \end{subfigure}
\setlength{\belowcaptionskip}{-15pt} 
\caption{Correlation matrices of student and teacher logits. * means correlation matrices of the normalized logits.}
\label{co_re}
\end{figure}

\section{Conclusion}
In this paper, we restudy the effect of hyper-parameter temperature and figure out the incapability of a single temperature to distill the full knowledge. To solve this issue, we propose a novel knowledge distillation method named NormKD which customizes the temperature for each sample according to the logits distribution. In our method, we set the standard variance of a sample's logit as its temperature, which helps the models transfer knowledge from each sample more equally. Furthermore, NormKD is simple and barely has extra cost. It can be a perfect complement to the vanilla KD, and also can be combined with other logit based methods easily for its simplicity. Especially, when it is applied to DKD, the performance of logit based method can improve further and precede the feature based method in more cases. It demonstrates the potential of logit based KD method. 

\section{Limitation and Future Work}
In our method, we assume the distribution of logit outputs as normal distribution and process the normalization operation to rearrange the outputs. However, the real situation is not so ideal and may make this assumption inappropriate. There may be a better method to be investigated in the future work, which can describe the logit distribution more correctly and help knowledge distillation to transfer knowledge from each sample more equally.

{\small
\bibliographystyle{ieee_fullname}
\bibliography{egbib}
}

\end{document}